\documentclass[conference]{IEEEtran}
\IEEEoverridecommandlockouts
\usepackage{cite}
\usepackage{amsmath,amssymb,amsfonts}
\usepackage{algorithmic}
\usepackage{graphicx}
\usepackage{textcomp}
\usepackage[table]{xcolor}
\usepackage{bm}
\usepackage{multirow}
\usepackage{colortbl}
\usepackage{floatrow}
\usepackage{makecell}
\usepackage{hyperref}

\hypersetup{urlcolor = blue}

\def\BibTeX{{\rm B\kern-.05em{\sc i\kern-.025em b}\kern-.08em
    T\kern-.1667em\lower.7ex\hbox{E}\kern-.125emX}}
\begin{document}

\def\eg{\emph{e.g.}} 
\def\Eg{\emph{E.g.}}
\def\ie{\emph{i.e.}} 
\def\Ie{\emph{I.e.}}
\def\cf{\emph{cf.}} 
\def\Cf{\emph{Cf.}}
\def\etc{\emph{etc.}} 
\def\vs{\emph{vs.}}
\def\wrt{w.r.t.} 
\def\dof{d.o.f.}
\def\iid{i.i.d.} 
\def\wolog{w.l.o.g.}
\def\etal{\emph{et al.}}
\makeatother

\title{
    Exploring Simple Siamese Network for High-Resolution Video Quality Assessment
}

\author{
\IEEEauthorblockN{Guotao Shen}
\IEEEauthorblockA{\textit{{\small Samsung \!Electronics \!(China) \!R\&D \!Centre}} \\
{\small Nanjing, China} \\
{\small guotao.shen@samsung.com}}
\and
\IEEEauthorblockN{Ziheng Yan}
\IEEEauthorblockA{\textit{{\small Samsung \!Electronics \!(China) \!R\&D \!Centre}} \\
{\small Nanjing, China} \\
{\small ziheng16.yan@samsung.com}}
\and
\IEEEauthorblockN{Xin Jin}
\IEEEauthorblockA{\textit{{\small Samsung \!Electronics \!(China) \!R\&D \!Centre}} \\
{\small Nanjing, China} \\
{\small xin.jin@samsung.com}}
\and
\IEEEauthorblockN{Longhai Wu}
\IEEEauthorblockA{\textit{{\small Samsung \!Electronics \!(China) \!R\&D \!Centre}} \\
{\small Nanjing, China} \\
{\small longhai.wu@samsung.com}}
\and
\IEEEauthorblockN{Jie Chen}
\IEEEauthorblockA{\textit{{\small Samsung \!Electronics \!(China) \!R\&D \!Centre}} \\
{\small Nanjing, China} \\
{\small ada.chen@samsung.com}}
\and
\IEEEauthorblockN{Ilhyun Cho}
\IEEEauthorblockA{\textit{Samsung Electronics}\\
{\small Suwon, Korea} \\
{\small ih429.cho@samsung.com}}
\and
\IEEEauthorblockN{Cheul-Hee Hahm}
\IEEEauthorblockA{\textit{Samsung Electronics}\\
{\small Suwon, Korea} \\
{\small chhahm@samsung.com}}
}

\maketitle

\begin{abstract}

In the research of video quality assessment (VQA), two-branch network~\cite{wu2023exploring} has emerged as a promising
solution. It decouples VQA with separate technical and aesthetic branches to measure the perception of low-level
distortions and high-level semantics respectively.  However, we argue that while technical and aesthetic perspectives
are complementary, the technical perspective itself should be measured in semantic-aware manner.  We hypothesize that
existing technical branch struggles to perceive the semantics of \textit{high-resolution} videos, as it is trained on
local mini-patches sampled from videos.  This issue can be hidden by apparently good results on low-resolution videos,
but indeed becomes critical for high-resolution VQA.  This work introduces SiamVQA, a simple but effective Siamese
network for high-resolution VQA.  SiamVQA shares weights between technical and aesthetic branches, enhancing the
semantic perception ability of technical branch to facilitate technical-quality representation learning.  Furthermore,
it integrates a dual cross-attention layer for fusing technical and aesthetic features.  SiamVQA achieves
state-of-the-art accuracy on high-resolution benchmarks, and competitive results on lower-resolution benchmarks.
Codes will be available at: \url{https://github.com/srcn-ivl/SiamVQA}

\end{abstract}


\section{Introduction}
\label{sec:intro}

Nowadays, smart mobile devices are ubiquitous. These devices are typically equipped with high-definition video
recording functionality, and produce high-resolution User Generated Content (UGC) videos every day. Therefore, it is of
significant value to measure the quality of these videos, on mobile devices where videos are captured, and on
social media platform where videos are uploaded, processed, and recommended.

In the last two decades, great efforts have been devoted to Video Quality Assessment (VQA), by gathering
enormous human quality
opinions~\cite{ying2021patch,gotz2021konvid,hosu2017konstanz,sinno2018large,wang2019youtube,wu2023exploring}, or
designing automatic VQA
models~\cite{wen2021strong,tu2020comparative,tu2021efficient,korhonen2019two,li2019quality,ying2021patch,li2022blindly,wu2022fast,wu2023neighbourhood,wu2023exploring}.
In recent years, deep learning based VQA models have developed into a successful
family~\cite{li2019quality,ying2021patch,wu2022fast,wu2023neighbourhood,wu2023exploring}, with compelling accuracy on
public benchmarks.  However, efficient and effective VQA remains challenging, due to the computational burden for
high-resolution videos and the diversity of contents of UGC videos.

\begin{figure*}[tb]
\centering
\includegraphics[width=0.88\textwidth]{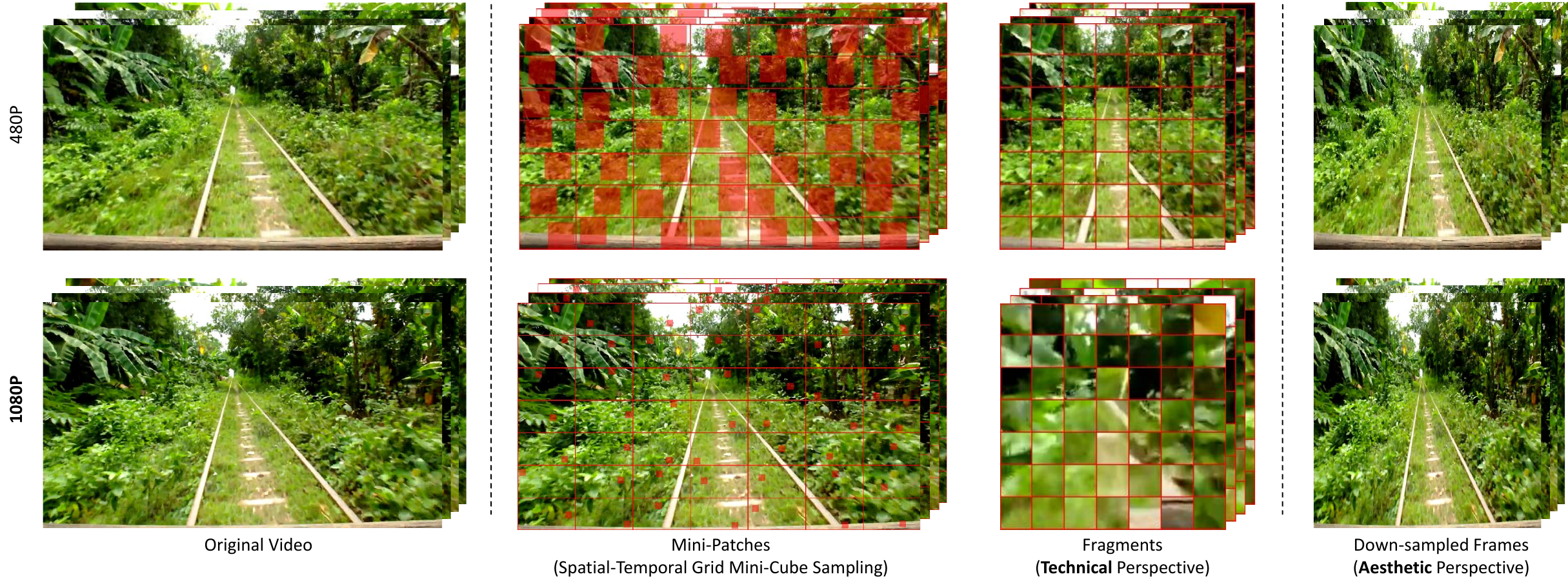}
\caption{Illustration of the fragments for technical perspective and down-sampled frames for aesthetic perspective.
We observe that the fragments sampled from \textit{high-resolution} videos (\eg, 1080p) suffer from serious
\textit{semantic degeneration}, although the fragments of lower-resolution videos can preserve the semantics to a large
extent.  In this example, even human can hardly tell the semantics of original 1080p video, purely based on the sampled
fragments.}
\label{fig:fragment}
\end{figure*}

To address the computational burden, Grid Mini-patch Sampling (GMS)~\cite{wu2022fast} and Spatial-temporal Grid
Mini-cube Sampling (St-GMS)~\cite{wu2023neighbourhood} are recently proposed to extract a set of local mini-patches from
original videos. Splicing these mini-patches forms fragments (see Fig.~\ref{fig:fragment}), a novel compact sample for
VQA that enables end-to-end representation learning at acceptable computational cost. Furthermore, since the
mini-patches are sampled at raw resolution, to a large extent, fragments can preserve technical quality cues that are
concerned with low-level distortions like blurs and artifacts.  However, the aesthetic perspective, which measures
semantic factors like contents and composition, may also affect human quality assessment for
videos~\cite{li2018has,gotz2021konvid}.  While fragments may contain rough scene-level
semantics~\cite{wu2023neighbourhood}, in Fig.~\ref{fig:fragment} we reveal the semantic degradation issue on
\textit{high-resolution} videos.

Taking both technical and aesthetic perspectives into account, DOVER~\cite{wu2023exploring} explicitly
decouples VQA with a two-branch network, with each branch measuring one perspective.  DOVER leverages fragments as
technical inputs, and down-sampled video frames as aesthetic inputs (see rightmost in Fig.~\ref{fig:fragment}).
Furthermore, DOVER employs inductive biases like branch architecture, pre-training and regularization to drive each
branch to focus on corresponding perspective, measuring two perspectives as separately as possible.

\begin{figure}[tb]
\centering
\includegraphics[width=0.88\columnwidth]{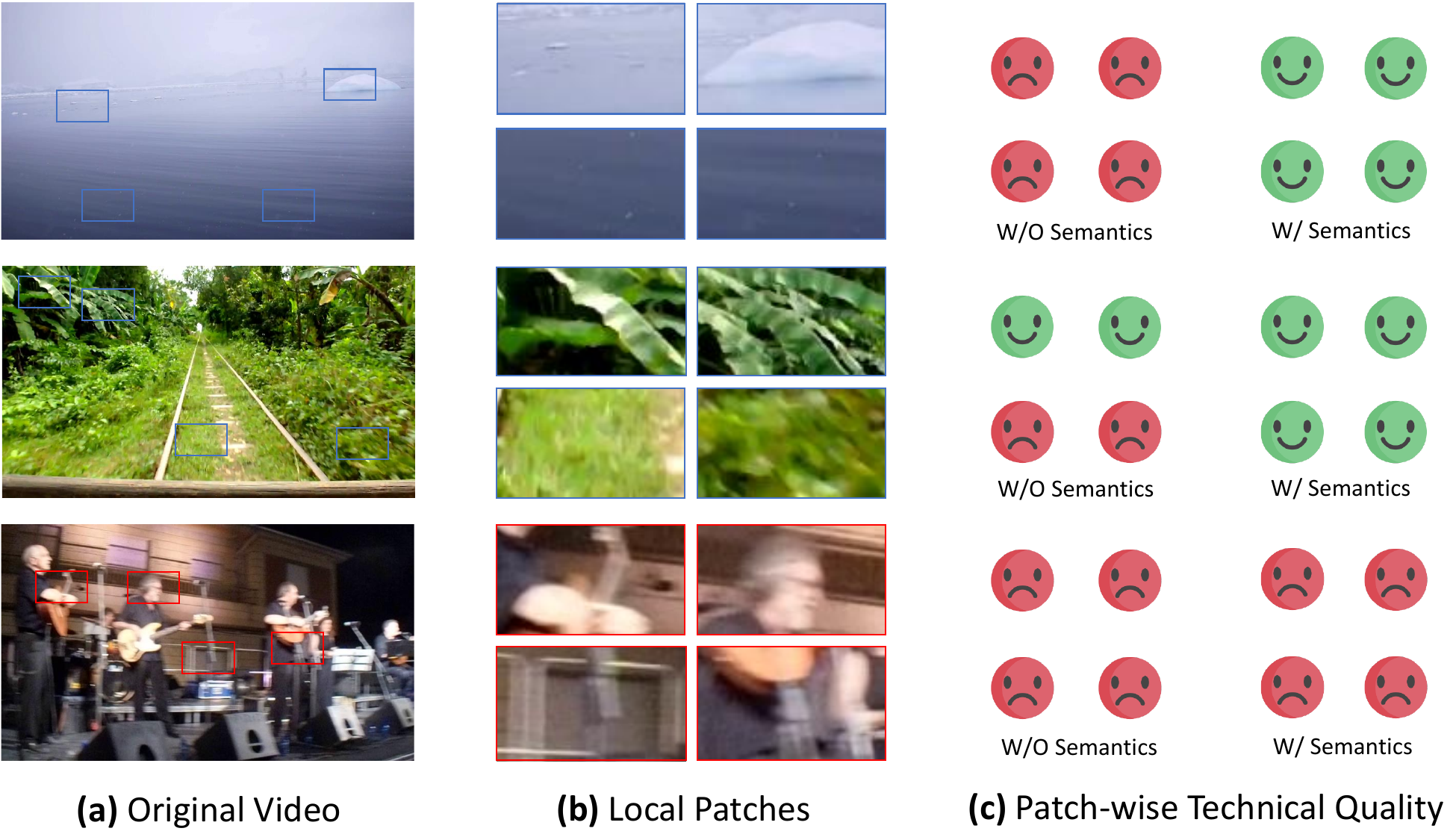}
\caption{Illustration of our argumentation that technical quality should be measured in context.
From first to third rows, we show videos of snowy scene, railway in forest with the camera lens moving forward fast, and
live show, respectively. Without considering the semantics, many local patches in first two videos are of low technical
quality (low lighting, motion blur). But with semantics, to a large extent these local patches are natural, because
heavy snow always leads to low-lighting, and through the lens of a fast-moving camera, nearby objects are always more
blurred than distant objects.  }
\label{fig:incontext}
\end{figure}

In this work, we challenge existing fragment-based technical quality evaluation for \textit{high-resolution} videos.  As
illustrated in Fig.~\ref{fig:incontext}, we argue that technical quality should be measured in context rather than
locally.  For example, a video of snow scene almost always has low-lighting, but such low-lighting does not indicate low
technical quality by considering the semantics. The technical branch in DOVER, which is separately trained on fragments,
may have limited ability to perceive semantics of high-resolution videos, resulting in inaccurate technical quality
scores for local mini-patches (see Fig~\ref{fig:quality-map} for examples).

Based on above analysis, we present SiamVQA -- a simple but effective Siamese network for high-resolution VQA. 
SiamVQA shares weights between aesthetic and technical branches, aiming to enhance the semantic perception ability of
the technical branch to boost technical-quality representation learning. Furthermore, SiamVQA employs a dual
cross-attention layer to fuse the high-level features of both branches.  We empirically show the effectiveness of the
design choices of SiamVQA.  SiamVQA achieves state-of-the-art accuracy on high-resolution public benchmarks, and
competitive results on lower-resolution benchmarks.

\section{Simple Siamese Network for VQA}
\label{sec:approach}

\subsection{Overview of SiamVQA Architecture}
\begin{figure}[tb]
\centering
\includegraphics[width=0.88\textwidth]{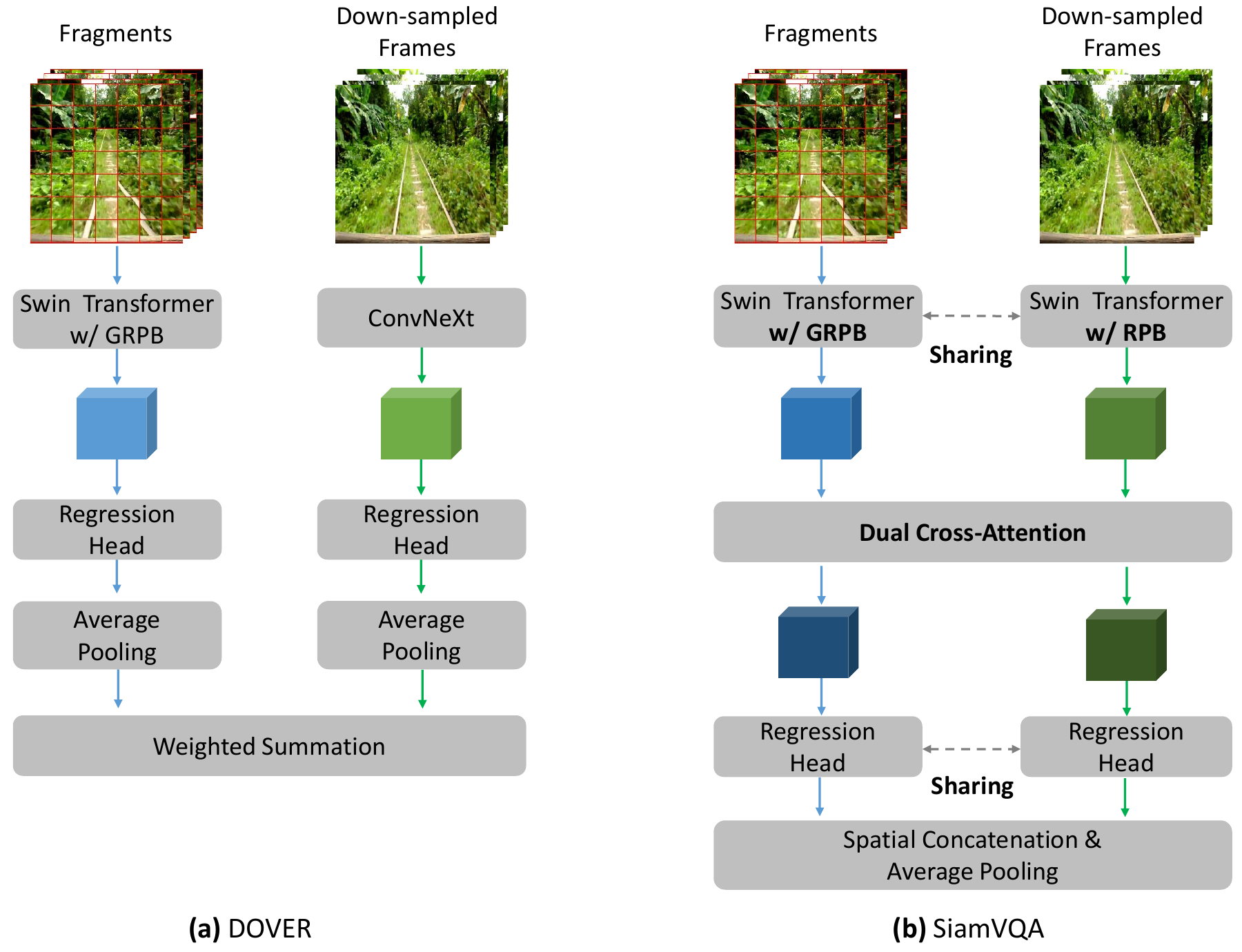}
\caption{Architecture of our SiamVQA, with comparison to DOVER~\cite{wu2023exploring}.}

\label{fig:arch}
\end{figure}

As shown in Fig.~\ref{fig:arch} (b), SiamVQA consists of two identity Swin-T~\cite{liu2021swin} for extracting both
technical and aesthetic features.  On top of Swin-T, a dual cross-attention layer is applied for feature fusion.  Then a
shared per-pixel regression head is employed to predict technical and aesthetic quality maps. Finally, these two quality
maps are concatenated and average pooled to produce the final prediction of video quality.

Formally, let $\mathbf{x}_t$ and $\mathbf{x}_a$ denote technical and aesthetic inputs,
$\mathcal{F}(\cdot)$ and $\mathcal{R}(\cdot)$ denote the backbone network and regression head,
$\{\mathbf{W}_t^Q, \mathbf{W}_t^K, \mathbf{W}_t^V\}$ and $\{\mathbf{W}_a^Q, \mathbf{W}_a^K, \mathbf{W}_a^V\}$ denote
the embedding weights of query, key, and value for the cross-attention layer of technical and aesthetic branches.
Then, technical and aesthetic quality maps $\mathbf{Q}_t$ and $\mathbf{Q}_a$, and the final quality score
$s$ can be calculated as:
{\small
\begin{align}\label{eq:quality-map}
    \mathbf{Q}_t & = \mathcal{R}(\mathrm{Attention}(\mathcal{F}(\mathbf{x}_t)\mathbf{W}_t^Q,
    \mathcal{F}(\mathbf{x}_a)\mathbf{W}_t^K, \mathcal{F}(\mathbf{x}_a)\mathbf{W}_t^V))  \nonumber\\
    \mathbf{Q}_a & = \mathcal{R}(\mathrm{Attention}(\mathcal{F}(\mathbf{x}_a)\mathbf{W}_a^Q,
    \mathcal{F}(\mathbf{x}_t)\mathbf{W}_a^K, \mathcal{F}(\mathbf{x}_t)\mathbf{W}_a^V))  \\
     s & = \mathrm{Pool}(\mathrm{Concat}(\mathbf{Q}_t, \mathbf{Q}_a)). \nonumber
\end{align}
}%

\subsection{Sharing Weights for Technical and Aesthetic Branches}
\label{subsec:share}
Weight-sharing in SiamVQA aims to improve technical-quality-related representation learning. This is based on our
realization that technical quality should be measured in context, and if the technical feature extractor itself is also
aesthetic feature extractor, it can perceive and leverage the semantics to boost technical-quality representation
learning.

Furthermore, we notice that semantic pre-training on Kinetics-400~\cite{kay2017kinetics} is crucial for
single-technical-branch VQA ~\cite{wu2023neighbourhood}, DOVER, and our SiamVQA. In analogy to semantic pre-training and
technical fine-tuning that performs sequentially in~\cite{wu2023neighbourhood}, our weight-sharing can be regarded as
semantic and technical training in parallel.  In Sec.~\ref{subsec:ablation}, we provide proof-of-concept experiment for
verifying. By removing semantic pre-training, we observe that our weight-sharing can improve SRCC on
LSVQ$_{1080p}$~\cite{ying2021patch} by as much as 5.6\%.

\subsection{Fusing Multimodal Features with Dual Cross-Attention}

We regard technical and aesthetic inputs as two modals for VQA. This is analogous to RGB video and optical flow for
action recognition, where feature fusion typically performs better than score
fusion~\cite{feichtenhofer2016convolutional,christoph2016spatiotemporal,feichtenhofer2017spatiotemporal,zolfaghari2017chained,gadzicki2020early}.
In SiamVQA, we employ dual cross-attention (see Eq.~\ref{eq:quality-map}) for feature fusion, aiming to mine useful
representations from complementary branches.

\subsection{Implementation Details}
\subsubsection{Technical and Aesthetic Inputs}
\label{para:inputs}
We generate fragments as in FasterVQA~\cite{wu2023neighbourhood}, and use down-sampled temporal frames as aesthetic
inputs. The input resolution of both our technical and aesthetic branches is $224\times224$.

\subsubsection{Position Encoding in SiamVQA}
We use the gated relative position bias (GRPB)~\cite{wu2022fast} for technical branch. While for aesthetic branch, we
use relative position bias (RPB), which is the default configuration in Swin Transformer~\cite{liu2021swin}.

\subsubsection{Loss Function}
We use a combined loss by weighting monotonicity loss $L_{mono}$ and differentiable PLCC loss $L_{plcc}$.
{\small
\begin{align}\label{eq:loss}
    L & = \lambda \cdot L_{mono} + L_{plcc}  \nonumber\\
    L_{mono} & = \sum_{i,j} \texttt{max}((\mathbf{s}_{pred}^{(i)} - \mathbf{s}_{pred}^{(j)})
    \texttt{sgn}(\mathbf{s}_{gt}^{(j)} - \mathbf{s}_{gt}^{(j)}), 0) \\
    L_{plcc} & = 1/2 \cdot (1 - \mathrm{PLCC}(\mathbf{s}_{pred}, \mathbf{s}_{gt})) \nonumber
\end{align}
}%
Here, $\texttt{sgn}(\cdot)$ is the sign function, $\mathbf{s}_{pred}$ and $\mathbf{s}_{gt}$ are label vectors from prediction and
ground truth in a mini-batch, $\mathbf{s}_{pred}^{(i)}$ and $\mathbf{s}_{gt}^{(i)}$ are the $i$-th scores in
$\mathbf{s}_{pred}$ and $\mathbf{s}_{gt}$.  $\lambda$ is set to 0.3.

\subsubsection{Training}

Swin-T in SiamVQA is initialized by pre-training on Kinetics-400~\cite{kay2017kinetics}.  SiamVQA is trained on
LSVQ$_{train}$~\cite{ying2021patch} consisting of 28,056 videos for 30 epochs.

\section{Experiments}
\label{sec:exp}

\subsection{Experimental Settings}
\label{exp:setting}

\subsubsection{Benchmark Datasets}
We report main results on high-resolution datasets, but also cover lower-resolution datasets.
\begin{itemize}
    \item High-resolution datasets: LSVQ$_{1080p}$~\cite{ying2021patch} is official test subset for LSVQ, which consists
        of 3,600 1080p videos. LIVE-Qualcomm~\cite{ghadiyaram2017capture} includes 208 videos of 1080p resolution, and
        YouTube-UGC~\cite{wang2019youtube} contains 1,380 videos with the max resolution of 4K.
    \item Lower-resolution datasets: We consider LSVQ$_{test}$~\cite{ying2021patch}, KoNViD-1k~\cite{hosu2017konstanz},
        and CVD2014~\cite{nuutinen2016cvd2014} as lower-resolution (from 240p to 720p) datasets for benchmarking.

\end{itemize}

\subsubsection{Evaluation Settings}
We evaluate VQA models under intra-dataset and transfer learning settings.
\begin{itemize}
    \item Intra-dataset evaluation: Since our model is trained on LSVQ$_{train}$, we consider the evaluation on
        LSVQ$_{test}$ and LSVQ$_{1080p}$ as intra-dataset evaluation.
    \item Transfer learning evaluation: To evaluate the transferability of learned representations, we fine-tune SiamVQA
        (trained on LSVQ$_{train}$) on smaller datasets (KoNViD-1k, CVD2014, LIVE-Qualcomm, and YouTube-UGC), and
        report transfer learning performance.
\end{itemize}

\subsubsection{Evaluation Metrics}
We use Spearman Rank-order Correlation Coefficient (SRCC) and  Pearson Linear Correlation Coefficient (PLCC) to evaluate
VQA accuracy. We measure the runtime of all models with a single A100 GPU.

\begin{table}[tb]
\centering
\footnotesize
% \scriptsize
\setlength{\tabcolsep}{0pt}
\begin{tabular*}{1.0\textwidth}{@{\extracolsep{\fill}}*{4}{lcc}}
\hline
\multirow{2}{*}{Experimental options} &  LSVQ$_{1080p}$ & Parameters\\
                         &  (SRCC$\uparrow$/PLCC$\uparrow$)  & (Million$\downarrow$) \\
\Xhline{2\arrayrulewidth}
single technical branch &  0.772/0.811  & 27.5  \\
single aesthetic branch  &  0.779/0.816  & 27.5  \\
two-branch, different structures (DOVER) &  0.795/0.830  & 55.6  \\
two-branch, same structure, unshared  &  0.797/0.837 & 56.0\\
\hline
two-branch, same structure, unshared &  0.797/0.837  & 56.0  \\
\quad \textbf{+ weight sharing} &  \textbf{0.803}/\textbf{0.841}  & \textbf{27.5}  \\
\quad + only technical inference &  0.601/0.640  & 56.0  \\
\quad \quad \textbf{+ weight sharing} &   \textbf{0.648}/\textbf{0.691} & \textbf{27.5}  \\
\quad + only aesthetic inference &  0.768/0.809  & 56.0  \\
\quad \quad + weight sharing &  0.773/0.809  & 27.5  \\
\quad + remove semantic pre-training &  0.629/0.685  & 56.0  \\
\quad \quad \textbf{+ weight sharing} &  \textbf{0.685}/\textbf{0.733}  & \textbf{27.5}  \\
\hline
two-branch, same structure, shared  &  0.803/0.841  & 27.5 \\
\quad + feature fusion: concat &  0.804/0.844  & 27.6  \\
\quad + feature fusion: self-attention &  0.803/0.841  & 32.3  \\
\quad \textbf{+ feature fusion: cross-attention} &  \textbf{0.805}/\textbf{0.848}  & 37.0  \\
\hline
\end{tabular*}
\caption{Ablation study of our design choices}
\label{tab:ablation}
\end{table}

\subsection{Ablation Study}
\label{subsec:ablation}

We conduct ablation experiments on LSVQ$_{1080p}$, and summarize the results in Tab.~\ref{tab:ablation}.

\subsubsection{A Simple Two-Branch Baseline}
We construct a simple two-branch baseline. It employs Swin-T for both branches without weight sharing, and directly
fuses technical and aesthetic scores for VQA.  Surprisingly, on 1080p videos, this simple baseline performs
competitively (even better) with DOVER which is based on different branch structures.

\subsubsection{Effectiveness of Weight-sharing}
We firstly show that our weight-sharing strategy improves over the unshared counterpart, with SRCC increased from 0.797
to 0.803.

We further investigate which branch contributes to the accuracy gain.  We note that with our two-branch baseline
trained, its technical or aesthetic branch can be used for prediction individually.  We observe that weight-sharing can
significantly improve SRCC from 0.601 to 0.648 when only using technical branch for inference, suggesting that our
overall accuracy gain mainly stems from the improved technical branch.

Furthermore, we remove the semantic pre-training on Kinetics-400, which significantly degrades the accuracy as verified
in~\cite{wu2023neighbourhood}. Under this setting, the improvement by weight-sharing is more significant, with SRCC from
0.629 to 0.685. This result confirms our second explanation in Sec.~\ref{subsec:share}.

\subsubsection{Effectiveness of Feature Fusion}
Our dual cross-attention strategy shows better accuracy than other feature fusion methods.  Cross-attention
tends to drive the network to learn more robust representations, by mining video-quality-related features from the
complementary branch.

\begin{figure*}[tb]
\centering
\includegraphics[width=0.88\textwidth]{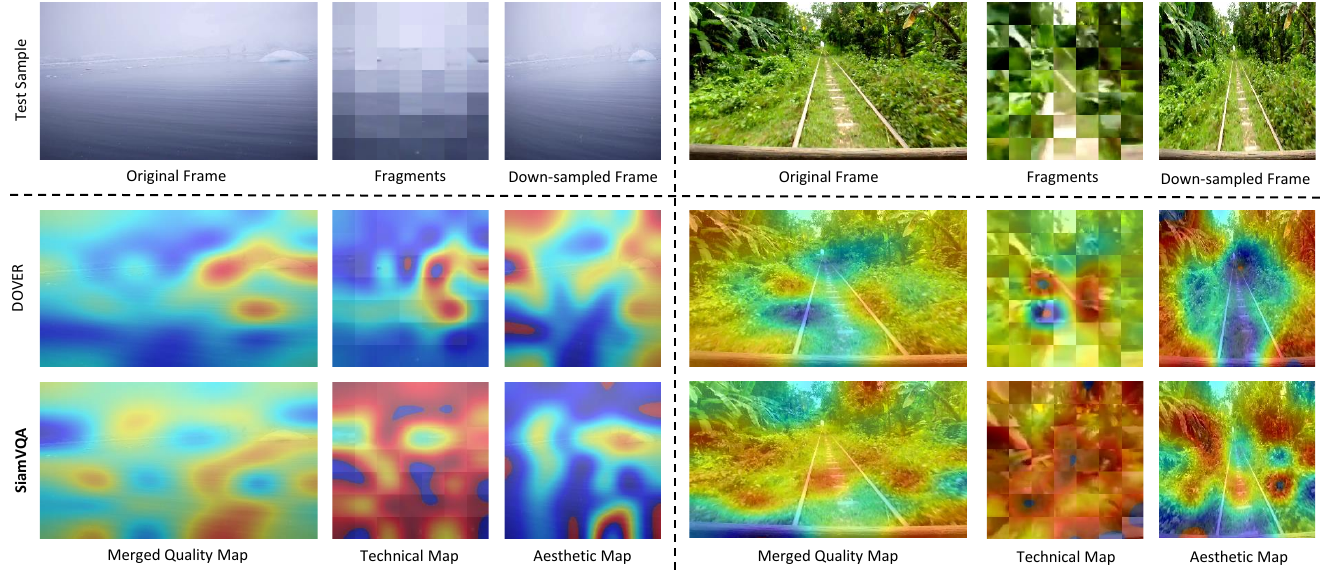}
\caption{Visualization of branch-level and merged quality maps, produced by DOVER~\cite{wu2023exploring} and our
    SiamVQA. For these two examples, SiamVQA gives  more higher technical scores on low-lighting and blur regions, as
    these low-level distortion appears in snowy scene, and video captured by fasting moving camera.
    The true quality scores of these two examples are 63.8 and 62.9; predictions by DOVER are 49.11 and 38.11;
    predictions by SiamVQA are 63.04 and 62.19.
}
\label{fig:quality-map}
\end{figure*}

\begin{table}[tb]
\centering
\footnotesize
% \scriptsize
\setlength{\tabcolsep}{0pt}
\begin{tabular*}{0.95\textwidth}{@{\extracolsep{\fill}}*{5}{lccc}}
\hline
\multirow{2}{*}{Methods} & Intra-dataset 
                         & \multicolumn{2}{c}{Transfer learning} \\
\cline{2-2}
\cline{3-4}
                         & LSVQ$_{1080p}$ & LIVE-Qualcomm & YouTube-UGC \\
\Xhline{2\arrayrulewidth}
VSFA~\cite{li2019quality}  & 0.675/0.704 & 0.737/0.732 & 0.724/0.743  \\
TCSVT~\cite{li2022blindly}  & 0.771/0.782 & 0.817/0.828 & 0.831/0.819  \\
FAST-VQA~\cite{wu2022fast}  & 0.779/0.814 & 0.819/0.851 & 0.855/0.852  \\
FasterVQA~\cite{wu2023neighbourhood}   & 0.772/0.811 & 0.826/0.844 & 0.863/0.859  \\
DOVER~\cite{wu2023exploring}  & 0.795/0.830 & 0.848/0.855 & 0.890/0.891  \\
\hline
SiamVQA (Ours) & \textbf{0.805}/\textbf{0.848} 
               & \textbf{0.867}/\textbf{0.886} 
               & \textbf{0.895}/\textbf{0.899} \\
\hline
\end{tabular*}
\caption{Qualitative (SRCC/PLCC) comparisons on \textbf{high-resolution} datasets under intra-dataset and transfer learning settings.}
\label{tab:highres}
\end{table}

\begin{table}[tb]
\centering
\footnotesize
% \scriptsize
\setlength{\tabcolsep}{0pt}
\begin{tabular*}{0.95\textwidth}{@{\extracolsep{\fill}}*{5}{lccc}}
\hline
\multirow{2}{*}{Methods} & Intra-dataset 
                         & \multicolumn{2}{c}{Transfer learning} \\
\cline{2-2}
\cline{3-4}
                         & LSVQ$_{test}$ & KoNViD-1k & CVD2014 \\
\Xhline{2\arrayrulewidth}
VSFA~\cite{li2019quality}  & 0.801/0.796 & 0.784/0.794 & 0.870/0.868  \\
TCSVT~\cite{li2022blindly}  & 0.852/0.855 & 0.834/0.837 & 0.872/0.869  \\
FAST-VQA~\cite{wu2022fast}  & 0.876/0.877 & 0.859/0.855 & 0.891/0.903  \\
FasterVQA~\cite{wu2023neighbourhood}   & 0.873/0.874 & 0.863/0.863 & \textbf{0.896}/0.904  \\
DOVER~\cite{wu2023exploring}  & 0.888/0.889 & 0.884/0.883 & 0.855/0.863  \\
\hline
SiamVQA (Ours) & \textbf{0.889}/\textbf{0.890} 
               & \textbf{0.887}/\textbf{0.888} 
               & 0.890/\textbf{0.907} \\
\hline
\end{tabular*}
\caption{Qualitative (SRCC/PLCC) comparisons on \textbf{lower-resolution} datasets under intra-dataset and transfer learning settings.}
\label{tab:lowres}
\end{table}

\subsection{Comparisons with State-of-the-art}

\subsubsection{On High-resolution Datasets}

As shown in Tab.~\ref{tab:highres}, our SiamVQA achieves excellent results on high-resolution LSVQ$_{1080p}$,
LIVE-Qualcomm, and YouTube-UGC datasets.  It outperforms DOVER~\cite{wu2023exploring} on all datasets, and surpasses all
others methods by a large margin.  In particular, when fine-tuned on LIVE-Qualcomm, SiamVQA outperforms DOVER by 1.9\% and
3.1\% in SRCC/PLCC. These results demonstrate the effectiveness of SiamVQA for high-resolution VQA.

\subsubsection{On Lower-resolution Datasets}

As shown in Tab.~\ref{tab:lowres}, SiamVQA also achieves competitive results on lower-resolution LSVQ$_{test}$,
KoNViD-1k, and CVD2014. It still outperforms DOVER on average, and show obvious advantages over other methods.  These
results suggest that our design choices towards high-resolution VQA do not sacrifice the performance on lower-resolution
VQA.

\begin{table}[tb]
\centering
\footnotesize
% \scriptsize
\setlength{\tabcolsep}{0pt}
\begin{tabular*}{0.95\textwidth}{@{\extracolsep{\fill}}*{7}{lccccc}}
\hline

\multirow{2}{*}{Methods} & \multirow{2}{*}{Tech.}
                         & \multirow{2}{*}{Aesth.}
                         & \multirow{2}{*}{Pre-train}
                         &  Param. & Runtime \\
                         & & &  & (Millions) & (Seconds) \\

\Xhline{2\arrayrulewidth}
FAST-VQA~\cite{wu2022fast} & \checkmark &  $\times$  & Kinetics-400   & 27.5 & 0.044   \\
FasterVQA\cite{wu2023neighbourhood} & \checkmark &  $\times$ & Kinetics-400  & 27.5  & 0.023    \\
DOVER~\cite{wu2023exploring}   &  \checkmark & \checkmark & Kinetics-400\&AVA & 55.6  & 0.065   \\
\hline
SiamVQA (Ours) & \checkmark & \checkmark & Kinetics-400 & 37.0 & 0.037 \\
\hline
\end{tabular*}
\caption{Comparisons in architecture, parameters, and runtime.}
\label{tab:param-runtime}
\end{table}

\subsection{Architecture, Parameters, and Runtime.}

Tab.~\ref{tab:param-runtime} gives the comparisons in architecture, parameters and runtime. Compared to DOVER, SiamVQA
has less parameters, runs faster, and does not require additional pre-training on aesthetic datasets (\eg.
AVA~\cite{murray2012ava}).

\subsection{Visualization of Quality Maps}

Fig.~\ref{fig:quality-map} shows the two examples of per-branch and merged quality maps, where SiamVQA can leverage
semantics to predict more accurate technical quality scores in context.

\section{Conclusion}
\label{sec:conclusion}

This work presented SiamVQA, a simple but effective Siamese network for high-resolution VQA.  It leverages
weight-sharing to enhance technical-quality-related representation learning in context, and achieve state-of-the-art
accuracy for high-resolution VQA.  We expect that our simple network design can encourage researchers to rethink the
design principles of two-branch VQA networks.

\bibliographystyle{IEEEtran}
\bibliography{siamvqa}

\end{document}